\documentclass[runningheads]{llncs}

\usepackage{xcolor}
\usepackage{times}
\usepackage{soul}
\usepackage{url}
\usepackage{amsmath}
\usepackage{hyperref}
\usepackage[nameinlink]{cleveref}
\usepackage[utf8]{inputenc}
\usepackage[small]{caption}
\usepackage{graphicx}
\usepackage{booktabs}
\usepackage{algorithm}
\usepackage{subcaption}
\usepackage{multirow}
\usepackage{subcaption}
\usepackage{float}
\usepackage{xspace}
\usepackage[normalem]{ulem}
\usepackage{amssymb}
\usepackage{algorithm}
\usepackage[noend]{algpseudocode}
\usepackage{listings}
\usepackage[nameinlink]{cleveref}
\usepackage{todonotes}

\lstdefinestyle{essence}{
 basicstyle=\ttfamily\scriptsize 
, keywords = {language, Essence, given, letting, find, such, that, where
, domain, function, total, surjective, be
, forAll, exists, injective, in, preImage, range
, int, mset, set, partition, new, type, intersect, matrix, from
, minimising, maximising, indexed, by, bool
, defined, minSize, maxSize, size, maxNumParts, numParts
, subset, subsetEq, relation
, toInt, sum
}
, frame = single
, framesep = 5pt
, morecomment=[l]{\$} 
}

\usepackage[nameinlink]{cleveref}

\newcommand{\essence}[0]{\textsc{Essence}\xspace}
\newcommand{\conjure}[0]{\textsc{Conjure}\xspace}
\newcommand{\eprime}[0]{\textsc{Essence Prime}\xspace}
\newcommand{\savilerow}[0]{\textsc{Savile Row}\xspace}
\newcommand{\minion}[0]{\textsc{Minion}\xspace}

\newcommand{\nbc}[0]{\textsc{nbc\_minisat\_all}\xspace}
\newcommand{\glucose}[0]{\textsc{glucose}\xspace}
\newcommand{\cadical}[0]{\textsc{cadical}\xspace}
\newcommand{\minisat}[0]{\textsc{minisat}\xspace}

\newcommand{\closed}[0]{\texttt{CFIS}\xspace}
\newcommand{\generator}[0]{\texttt{GFIS}\xspace}
\newcommand{\minimal}[0]{\texttt{MRIM}\xspace}
\newcommand{\discriminative}[0]{\texttt{DFIM}\xspace}
\newcommand{\relevantsub}[0]{\texttt{RSD}\xspace}

\newcommand{\hide}[1]{}

\usepackage{tikz}
\usetikzlibrary{shapes.geometric, arrows, positioning}
\tikzstyle{startstop} = [rectangle, rounded corners, minimum width=2cm, minimum height=1cm,text centered, draw=black]
\tikzstyle{io} = [trapezium, trapezium left angle=70, trapezium right angle=110, minimum width=2cm, minimum height=1cm, text centered, draw=black]
\tikzstyle{process} = [rectangle, minimum width=35mm, minimum height=1cm, text centered, text width=38mm, draw=black]
\tikzstyle{decision} = [diamond, minimum width=2cm, minimum height=1cm, text centered, draw=black]
\tikzstyle{arrow} = [thick,->,>=stealth]

\begin{document}

\title{Efficient Incremental Modelling and Solving}

 \author{Gökberk Koçak, Özgür Akgün, Nguyen Dang, Ian Miguel}
 \authorrunning{G Koçak et al.}

 \institute{
 School of Computer Science, University of St Andrews, UK\\
 \email{\{gk34,ozgur.akgun,nttd,ijm\}@st-andrews.ac.uk}
 }

\maketitle

\begin{abstract}
In various scenarios, a single phase of modelling and solving is either not sufficient or not feasible to solve the problem at hand. A standard approach to solving AI planning problems, for example, is to incrementally extend the planning horizon and solve the problem of trying to find a plan of a particular length. Indeed, any optimization problem can be solved as a sequence of decision problems in which the objective value is incrementally updated. Another example is constraint dominance programming (CDP), in which search is organized into a sequence of levels. The contribution of this work is to enable a native interaction between SAT solvers and the automated modelling system \savilerow to support efficient incremental modelling and solving. This allows adding new decision variables, posting new constraints and removing existing constraints (via assumptions) between incremental steps. Two additional benefits of the native coupling of modelling and solving are the ability to retain learned information between SAT solver calls and to enable SAT assumptions, further improving flexibility and efficiency. Experiments on one optimisation problem and five pattern mining tasks demonstrate that the native interaction between the modelling system and SAT solver consistently improves performance significantly.

\end{abstract}

\section{Introduction}

When approaching the solution of a class of problems, in many cases a simple single-phase approach works well: formulate a model parameterised on the data that defines an individual instance of the problem class, and solve each instance in a single solving phase. In some scenarios however, as we will illustrate below, this approach is either not sufficient or not feasible to solve the problem at hand. Instead, a larger or more difficult problem instance is solved as a sequence of smaller or simpler related instances. In this situation, communication between a modelling system that prepares an instance for solution for a low-level solver and the solver itself can become a bottleneck, with much work repeated between consecutive, very similar instances.

Incremental modelling and solving is a process of constructing an initial low level instance and
obtaining further instances in a sequence by modelling and encoding just the differences between the previous and the new instance. Most SAT solvers are capable of working incrementally by allowing to append new irrevocable clauses or set certain assumptions that are temporary to each call. 

To illustrate, consider the task of pattern mining, the process of extracting useful patterns from large data sets. The most well-known pattern mining task, frequent itemset mining \cite{agrawal1994fast}, requires us to find the sets of items whose number of occurrences together (known as the {\em support}) in a transactional database exceeds a specified threshold. Specialised, efficient tools exist for standard pattern mining tasks \cite{zaki2000scalable}. However, finding all frequent patterns is rarely useful since it usually produces a very large volume of results. Rather, an end-user is typically interested in focusing on a much smaller set of patterns for further inspection. One approach is to seek patterns that compactly represent the full set of patterns \cite{pasquier1999discovering}, another is to consider domain-specific side constraints \cite{bonchi2004closed} that further reduce the volume of patterns returned. Both methods require a more sophisticated search for patterns and hence carry an increase in computational cost.

Constraint-based mining \cite{de2008constraint} offers a general means of modelling more sophisticated pattern mining tasks. Its flexibility means that side constraints can easily be added to the basic model of a pattern mining problem, which is difficult to do with a specialised mining tool. We distinguish {\em local} and {\em non-local} constraints in modelling pattern mining problems. The former, such as the frequent itemset property, can be expressed simply on a candidate solution, e.g. by constraining the support of a candidate itemset to be equal to or greater than the threshold. Non-local constraints, however, must be expressed {\em between} candidate solutions and are therefore more challenging to model. Closed frequent itemset mining \cite{pasquier1999discovering,koccak2018closed}, which is one approach to representing the full set of frequent itemsets more compactly, is an illustrative example: it stipulates that an itemset is closed frequent if its support exceeds that of all of its supersets.

Constraint Dominance Programming (CDP) \cite{negrevergne2013dominance} provides a method of supporting constraints between solutions via {\em dominance blocking constraints}: every time a new solution is found, a new blocking constraint is added to disallow solutions that it would dominate. An extension to CDP, CDP+I \cite{koccak2019towards,kocak2020} exploits {\em incomparability} between solutions (solutions $A$ and $B$ are incomparable if $A$ does not dominate $B$ and $B$ does not dominate $A$) so that they may be found in batches. The search is organized into levels in which all solutions are incomparable, and hence may be found together through a single call to a solver without the need for additional per-solution blocking constraints. Operating on CDP, which requires posting new constraints after each solution, and operating on CDP+I, which has requirements similar to CDP's but for a batch of solutions, are incremental modelling and solving examples.

Other problem types that might be considered for incrementality are constrained optimisation problems (COP), where an objective function is given in addition to a standard constraint satisfaction problem, or AI planning problems where we can incrementally extend the planning horizon and solve the problem of trying to find a plan of a particular length.

CP solvers like \minion~\cite{gent2006minion} or chuffed~\cite{chuchuffed} are typically capable of supporting COP directly in addition to CSP. However other solver types, such as standard SAT solvers, sometimes lack the facility to represent objective values. Instead of using standard SAT encoding for the problem, a maximal satisfiability problem encoding (MaxSAT) can be used to represent the objective function. However, converting a SAT encoding to a MaxSAT encoding may be time consuming depending on the size of the instance.

Alternatively, using SAT or SMT solvers is possible for optimisation and planning problems via a sequence of solver calls in an incremental structure. The COP can be encoded as a pair of CSP's with a different optimisation value encoded into each CSP. Afterwards, those CSP instances can be solved for satisfiability. The threshold where the problem switches from SAT to UNSAT or the other way around can indicate the proven optimal value for the original COP instance. Searching for this threshold will have multiple solver calls that can be adjusted for efficiency.

\subsubsection{Contribution} This paper proposes to enable a native interaction between the SAT solver and the automated modelling system that organizes the CDP+I mining process and the optimization process using a SAT backend. This is done to remove a major bottleneck in which the consecutive SAT calls are operated. Two additional benefits of this native coupling are the ability to retain learned information between SAT solver calls and to enable SAT assumptions, further improving efficiency by reducing redundant search between levels. 

Our experiments on one optimization problem and five pattern mining tasks demonstrate that the native interaction between the modelling system and SAT solver consistently improves the performance of each system significantly.

\section{Background}
\subsection{CDP+I Architecture}
\essence~\cite{Akgun2013automated} is an abstract high-level constraint specification language. It has the power to represent complex abstract structures, such as sets, multisets, sequences, and partitions. It supports arbitrary nesting of these structures and also supports quantification over decision variables. Hence, the language is ideally suited to expressing data mining problems. \essence can be refined into a constraint model in \eprime~\cite{essence-prime-description} using \conjure~\cite{Akgun2013automated}. Due to the high-level abstract nature of the specification, there are multiple ways of compiling \essence to \eprime. \conjure has a number of built in heuristics to make modelling decisions automatically. Alternatively, the modelling decisions can be manually selected. \savilerow translates \eprime{} into input suitable for a variety of black-box solvers while applying solver specific optimisations to the model, such as rewriting constraint expressions, common sub-expression elimination and using \minion{} to enforce strong levels of consistency in a preprocessing step \cite{nightingale2015automatically}.

A constraint satisfaction problem consists of decision variables ($V$), their domains ($D$) and problem constraints ($C$). CDP extends constraint satisfaction problems (CSP) by adding a \textit{dominance relation} ($R$), which defines the condition under which an assignment to the decision variables is dominated by another assignment. In CDP, an assignment is a solution if it is not dominated by any other solution. When enumerating all solutions of a CDP instance, dominance blocking constraints can be generated for each solution as soon as they are found. These constraints will eliminate all future dominated assignments. However, a post-processing step may still be needed~\cite{negrevergne2013dominance}. CDP+I extends CDP by defining an \textit{incomparability function} ($I$), which defines when two assignments are incomparable (mutually non-dominating).

\begin{figure}[t!]
\centering
\linespread{1.3}
\begin{lstlisting}
language Essence 1.3
letting ITEM be domain int(...)
letting SUPPORT be domain int(...)
given db : mset of set of ITEM
given minSupport : int
find itemset: set of ITEM
find support: SUPPORT
such that
 support = sum entry in db . toInt(itemset subsetEq entry),
 support >= minSupport,
 SideConstraints
 \end{lstlisting}
 \begin{lstlisting}
dominanceRelation 
  (itemset subsetEq fromSolution(itemset))
  -> (support != fromSolution(support))
   \end{lstlisting}
 \begin{lstlisting}
incomparabilityFunction descending |itemset|
 \end{lstlisting}
 \caption{Closed Frequent Itemset Mining in \essence{}\label{fig:essenceClosed}. The dominance relation defines the closedness property between the currently sought solution and the previous solutions via \texttt{fromSolution}. The incomparability function is defined on cardinality using a descending order, since closedness is defined by a superset relation.}
\end{figure}

An itemset mining problem can be specified naturally in \essence as a multiset of transactions. Depending on the nature of the mining task, each transaction can be represented using a set of integer item labels or ornamented (using tuples or records) with additional information such as a class label. \Cref{fig:essenceClosed} presents the specification of the Closed Frequent Itemset Mining problem in three parts. The first part is the declaration of the parameters, the decision variables and any constraints that concern a single solution. The second part gives the dominance relation in terms of previously found solutions. The third part defines the incomparability function, which in this problem is any two solutions that have the same itemset cardinality.

\begin{algorithm}[h!]
\caption{CDP+I}\label{al:cdpi}
\begin{algorithmic}[1]
\State $ (V,D,C,R,I) \gets CDP\text{+}I $
\State $ levels \gets getLevels(I) $
\For{$ l \gets levels $}
    \State $ C \gets C \cup levelRestriction(l) $
    \State $ CSP \gets (V,D,C) $
    \State $ S \gets findAllSolutions(CSP) $
    \State $ B \gets generateDominanceBlocking(R,S) $
    \State $ C \gets C - levelRestriction(l) $
    \State $ C \gets C \cup B $
\EndFor
\end{algorithmic}
\end{algorithm}

\Cref{al:cdpi} makes use of both the dominance relation and the incomparability function when solving CDP+I instances. The CDP+I algorithm aims to find all non-dominated solutions. It achieves this by partitioning the search space into levels extracted from the incomparability function. For example, for the closed itemset mining problem, a separate search is conducted for every value in the domain of \texttt{|itemset|}. For every level, we take the base CSP model and start by adding a level restriction constraint to it. In our running example, this corresponds to posting a cardinality constraint on the itemset. Then, we enumerate all solutions and generate the corresponding dominance blocking constraints. The problem constraints are then updated to \textit{remove} the level restriction constraint before adding the new dominance blocking constraints.

Previous implementations of CDP+I made a separate solver call for each level when using an AllSAT solver and a separate solver call for each solution when using a standard SAT solver. This allows for a simple implementation of the CDP+I algorithm at the cost of losing learned clauses between separate solver calls. The performance of modern SAT solvers relies heavily on learned clauses~\cite{marques2009conflict}. \Cref{sec:native} presents our approach for enabling native interaction with SAT and AllSAT solvers. Through the use of assumptions in SAT, we achieve improved performance without changing the high-level problem specifications.

The use of \essence{} for specifying the problems allows access to a large number of different models (via \conjure{} options), different preprocessing options (via \savilerow{} options), and different solvers (SAT and AllSAT).
\subsection{Solving COP using SAT solvers}
\label{sec:cop_strategies}

A COP problem can be rewritten as a series of CSP problems where the objective function value is encoded differently in each of them. A naive but inefficient approach would be to exhaustively try all possible values and pick the best one which satisfies the instance. 
Alternatively, we can apply a search for the optimal objective function value in its domain space. Three different search strategies which are supported by \savilerow can be considered for this purpose, namely Linear, UNSAT, and Bisect. They are explained as follows (assuming that we are solving a maximisation problem).

\subsubsection{Linear search}
Linear search is a straightforward search strategy to search for the optimal value. It starts from the lowest value and increase the optimal by one incrementally until the problem becomes unsatisfiable.

\subsubsection{UNSAT search} This is also a straightforward strategy which starts from the highest objective function value and decreases it one by one until the problem becomes satisfiable. %

\subsubsection{Bisect search}
This is a binary search strategy also known as \emph{dichotomic} search. It starts with splitting the objective function's domain into two. This results in two CSP problems, each with half of split domain. The satisfiable CSP problem is chosen and the same procedure is repeated until the objective function's domain size reduce to one (the optimal objective function value).

\section{Problem Classes and Side Constraints}
\label{sec:Classes}

Throughout this paper we will experiment on six problem classes to demonstrate the enhancements we will introduce. Five of these problem classes are pattern mining problems encoded in CDP+I and the instances we use are taken from the supplementary material of \cite{kocak2020}. The sixth problem class is Multi-Mode Resource Constrained Project Scheduling Problem (MRCPSP). 

The pattern mining problems are variations of the frequent itemset mining problem, each parameterised over a dataset of transactions. The task is to find a set of frequent items that satisfy minimum value and maximum cost side constraints. In addition, each problem class has a different constraint among assignments which encodes the dominance relationship.

\subsubsection{Closed frequent itemset mining (CFIS)}
A frequent itemset is \textit{closed} if and only if its support is greater than all of its supersets \cite{pasquier1999discovering}. The \textit{support} of an itemset is the number of times the set occurs together in the transactions database. Maximal itemset mining is a similar problem class where the only difference is that a frequent itemset is \textit{maximal} if none of its supersets are frequent. We do not include maximal itemset mining in our experiments since it is a simpler version of closed itemset mining.
\subsubsection{Generator frequent itemset mining (GFIS)}
Generator itemsets (also called free itemsets or key itemsets) \cite{boulicaut2000approximation} are frequent itemsets which do not have any frequent subsets with the same support.
\subsubsection{Minimal rare itemset mining (MRIM)}
A minimal rare itemset is an infrequent itemset whose subsets are all frequent \cite{szathmary2007towards}.
\subsubsection{Closed discriminative itemset mining (DFIS)}
Discriminative itemset mining \cite{cheng2007discriminative} is parameterised over a dataset of transactions that also have a class label (positive/negative). Instead of a single support value, we maintain two support values: the \textit{positive support} of an itemset is the number of transactions that are labelled positive and have the itemset as a subset. The \textit{negative support} similarly is the number of transactions that are labelled negative and have the itemset as a subset. A \textit{discriminative itemset} is one where the difference between the positive and the negative support is greater than a given threshold. A closed discriminative itemset is a discriminative itemset that has support greater than all of its supersets.
\subsubsection{Relevant subgroup discovery (RSD)}

Relevant subgroup discovery~\cite{lemmerich2010fast} is similar to discriminative itemset mining. While discriminative itemset mining reasons on the support numbers of different classes of transactions, relevant subgroup discovery reasons using the actual sets of transactions that provide the support~\cite{negrevergne2013dominance}. A relevant subgroup $X$ is an itemset where at least one of following conditions hold; 1) For positive transactions, no other itemset covers a superset of the transactions covered by $X$, 2) For negative transactions, no other itemset covers a subset of the transactions covered by $X$ or 3) For both kinds of transactions, no other itemset that has the same total cover is a superset of $X$.

\subsubsection{Multi-mode resource constrained project scheduling problem (MRCPSP)} 
This is a variant of the project scheduling problem~\cite{kolisch1997psplib}, a classical and well-known optimisation problem in operations research. Given a number of activities and a set of renewable resources. Each activity is associated with a duration and demands for some resources. The activites are non-interrupted and there are precedence constraints which states that some activities can only start once some others are finished. The variant considered in this paper is the \emph{multi-mode}~\cite{mori1997genetic}, where each activity may have multple modes. Each mode dictates the duration and resource demands of the activity. The goal is to schedule the activities and choose a mode for each of them so that the makespan (the latest completion time) is minimised. An \essence  specification of this problem is presented in \Cref{sec:ap_mrcpsp} (\Cref{fig:mrcpsp}).

\section{Native Interaction} \label{sec:native}

The main CDP+I algorithm (\Cref{al:cdpi}) and the SAT optimisation backend requires multiple solver calls. For CDP+I, each solver calls occur once per level when using an AllSAT solver and once per solution when using a standard SAT solver. Solutions from a level are used to produce dominance blocking constraints for the next level. Furthermore, level restriction constraints are both added and removed between levels. Likewise, for optimisation problems using a standard SAT backend, multiple solver calls occur to apply three optimization strategies to reach to optimal value. In addition to adding temporary constraints, the ability to remove added constraints is also required.
Adding constraints during search is relatively common even without an incremental process. However, removing constraints requires special treatment by the solver in question. A direct implementation of these algorithms would indeed call the solver several times and consequently would not benefit from any learned clauses between solver calls.

There are two main ways of maintaining learned clauses between solver calls. The first option works by extracting learned clauses once the solver finishes the search and post-processing them to keep a relevant subset for a future solver invocation. \cite{shishmarev2016learning} uses a similar approach to learn candidate implied constraints from a learning solver. The second option works by keeping the solver active, modifying the active model by posting additional constraints and restarting search. Adding new variables and constraints in this way is a relatively common operation, available in \texttt{ipasir}, an incrementality API for SAT solvers used in SAT competitions \cite{jarvisalo2012international}. Removing constraints requires the {\em assumptions} machinery that is available in most modern SAT solvers. Constraints that are going to be removed are posted as conditional new clauses dependent on an assumption. Hence, when the assumption is lifted (and the constraint is removed) any learned clauses which depend on that assumption can be deactivated.

We define a new API for SAT solvers that shares most of the functionality of \texttt{ipasir}, including methods for adding new clauses, adding assumptions, solving and retrieving solutions. We extend this basic API to also include methods for reporting detailed statistics about learned clauses and the solver's state, in addition to triggering solution callbacks. Our extended API is implemented using the Rust programming language. It works with SAT solvers \glucose, \cadical and \minisat and the AllSAT solver \nbc. Our Rust implementation encapsulates the required functionality of these solvers and compiles them into a shared library.

The entire pipeline of tools starts with \conjure{}, which produces an \eprime{} model for each problem class. A modified \savilerow{} is then used to instantiate the problem class model using a given data file, preprocessing it using \minion{} to shave domains, and then encoding into SAT using the standard encodings found in \savilerow{} \cite{nightingale2017automatically}. Prior to our work, \savilerow{} worked by producing a DIMACS file that has the entire encoding in it and calling a SAT solver on this file. Thanks to the new API we define and implement, \savilerow now skips building this file and directly makes calls to the SAT solver to create the model.

Our solver API layer is implemented in Rust while \savilerow{} was implemented in Java. We use the Java Native Interface (JNI) to integrate the API layer into \savilerow{}.

\section{Experiments}

\subsection{MRCPSP experiments}
\label{sec:mrcpsp_experiments}

To demonstrate the effectiveness of keeping SAT learnt clauses between levels during the optimisation process using native interaction, we evaluate the three optimisation strategies explained in \Cref{sec:cop_strategies} on 928 MRCPSP instances from the PSPlib~\cite{kolisch1997psplib}. The SAT solver \glucose~\cite{audemard2018glucose} is combined with each of the three optimisation strategies. We also compare the the resulting performance with Open-WBO~\cite{martins2014open}, a MaxSAT solver and with \texttt{Chuffed}~\cite{chuchuffed}, a learning CP solver.

\begin{figure*}[!h]
    \centering
    \begin{subfigure}{0.3\textwidth}
    \includegraphics[width=\textwidth]{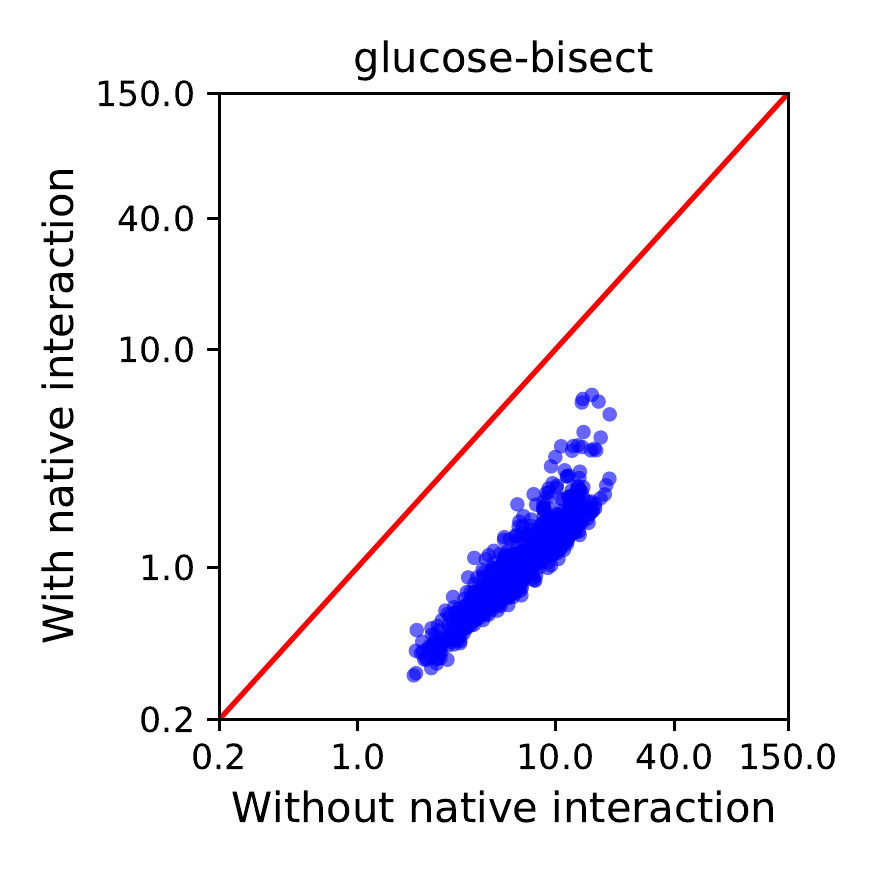}
    \end{subfigure}
    \begin{subfigure}{0.3\textwidth}
    \includegraphics[width=\textwidth]{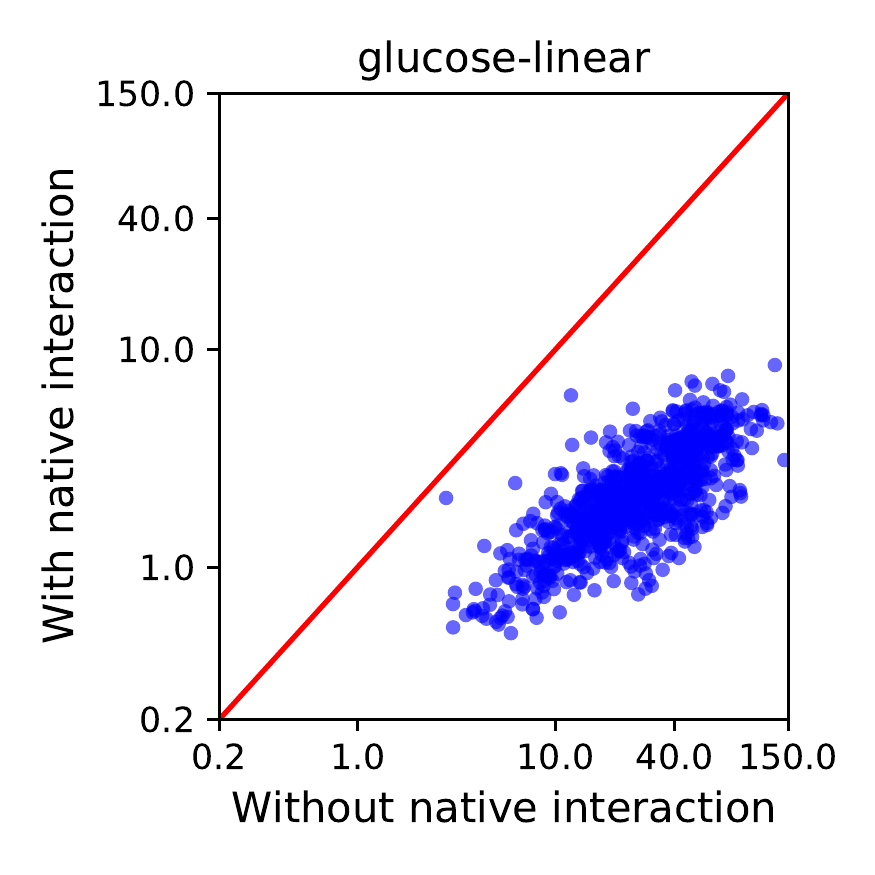}
    \end{subfigure}
    \begin{subfigure}{0.3\textwidth}
    \includegraphics[width=\textwidth]{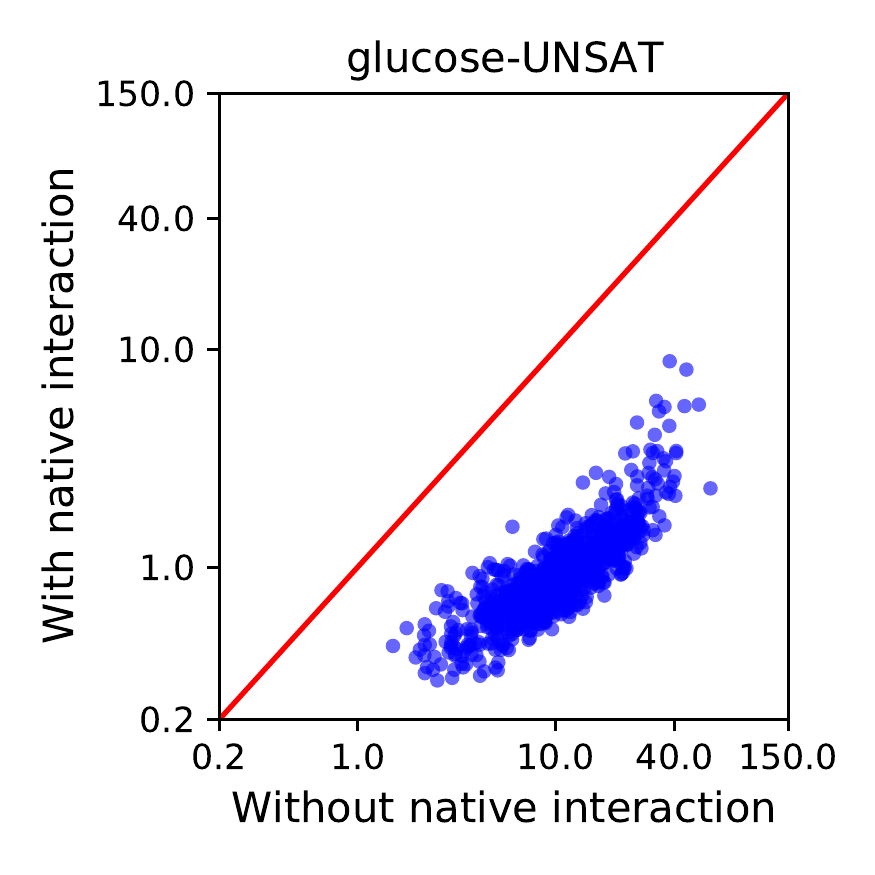}
    \end{subfigure}
    \caption{Solving time of \glucose with versus without native interaction on 928 MRCPSP instances.}
    \label{fig:inter_compare_mrcps}
\end{figure*}

Each run on an instance is given a time limit of one CPU hour, and is repeated three times. The average solving time is recorded. 
The comparison of the usage of native interaction on \glucose is shown in \Cref{fig:inter_compare_mrcps}. Results suggest that for all three strategies, the native interaction boosts the efficiency significantly on all tested instances.

\begin{figure*}[!h]
    \centering
    \begin{subfigure}{0.8\textwidth}
    \includegraphics[width=\textwidth]{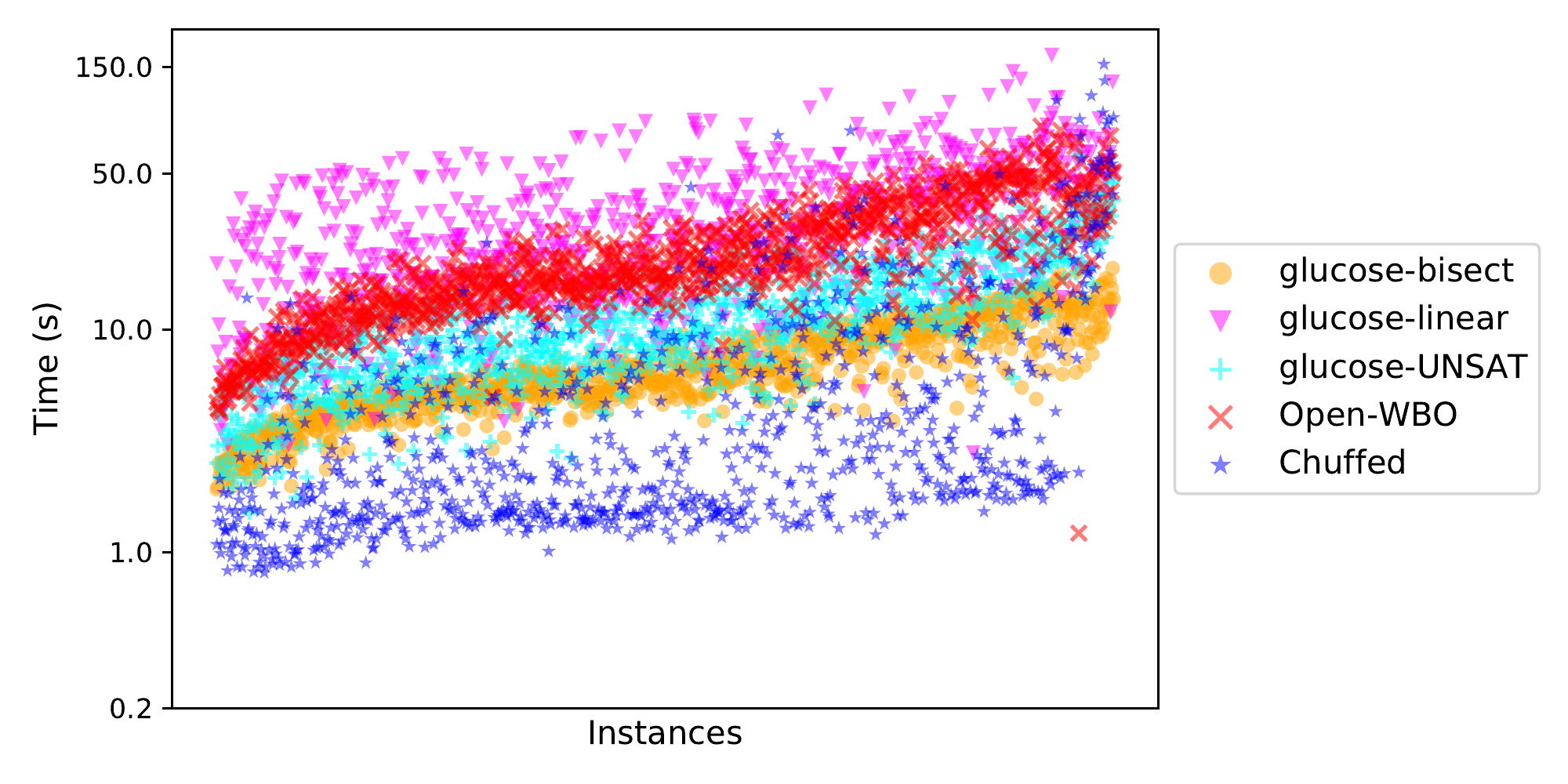}
    \caption{Without native interaction}
    \end{subfigure}
    \begin{subfigure}{0.8\textwidth}
    \includegraphics[width=\textwidth]{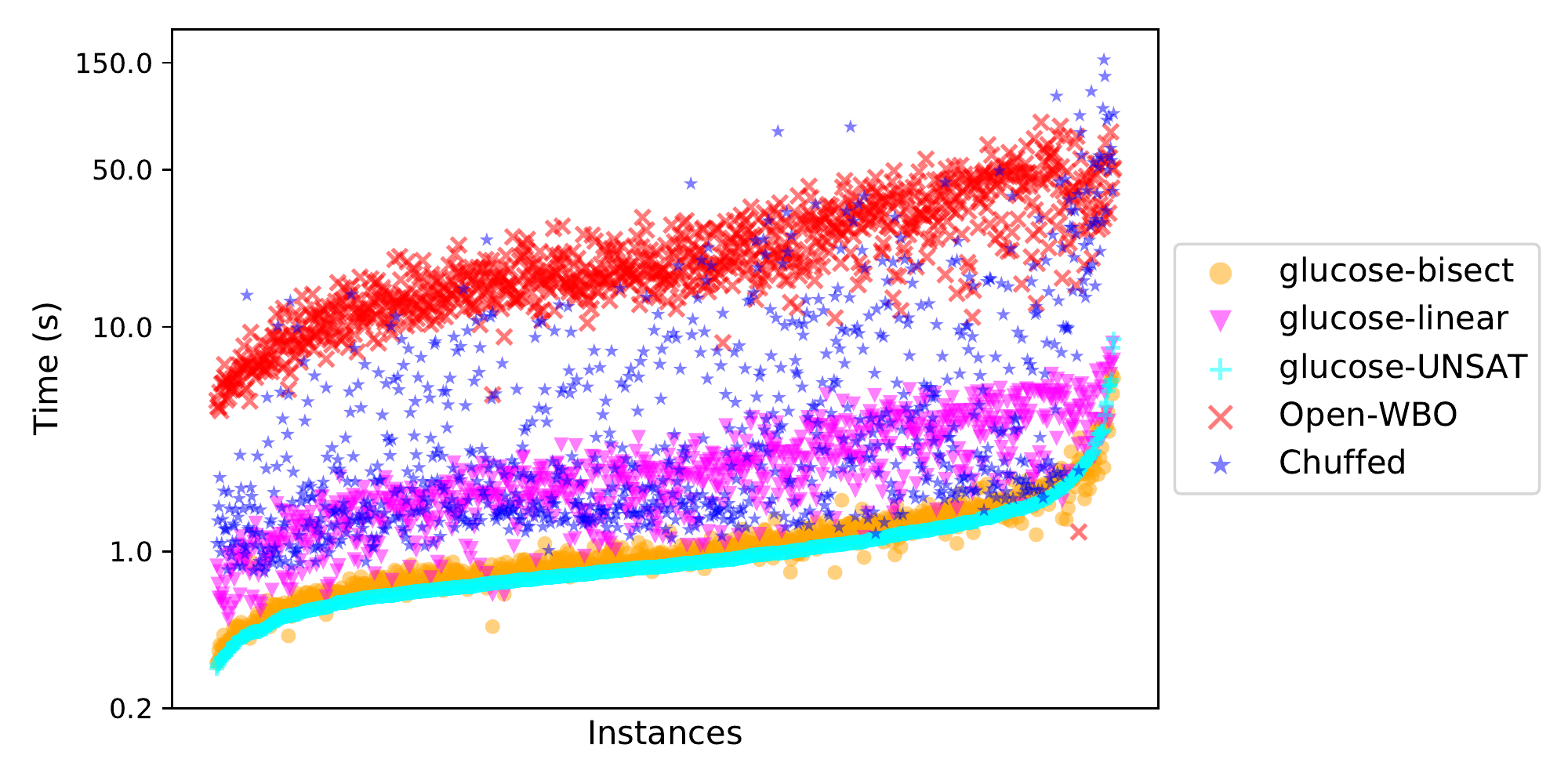}
    \caption{With native interaction}
    \end{subfigure}
    \caption{Solving time of \glucose with three settings (bisect, linear and UNSAT), Open-WBO and Chuffed on 928 MRCPSP instances. \glucose's results are shown without (top) and with (bottom) native interaction.}
    \label{fig:inter_mrcpsp}
\end{figure*}

Comparison against Open-WBO and Chuffed are plotted in \Cref{fig:inter_mrcpsp}. While in the first figure only includes the default SAT strategies, the second figure replaces them with their native equivalents.  Results suggest that the native interaction create a drastic performance improvement for the SAT backend \glucose and results on these problem instances are competitive against the two established optimisation solvers.

\subsection{CDP+I experiments}

\subsubsection{Computational Evaluation with a Standard SAT Solver}

In order to evaluate the effectiveness of maintaining learned clauses and using SAT assumptions between CDP+I levels, we solve 240 instances across 5 problem classes (see Section \ref{sec:Classes}). Within a 6-hour time limit, the native version solves 210 instances whereas pure CDP+I solves only 173 instances. We believe this is due to needing fewer search nodes, which is made possible by pruning large parts of the search tree via the learned clauses.

\begin{figure*}[h!]
    \begin{subfigure}{.5\textwidth}
        \includegraphics[width=\textwidth]{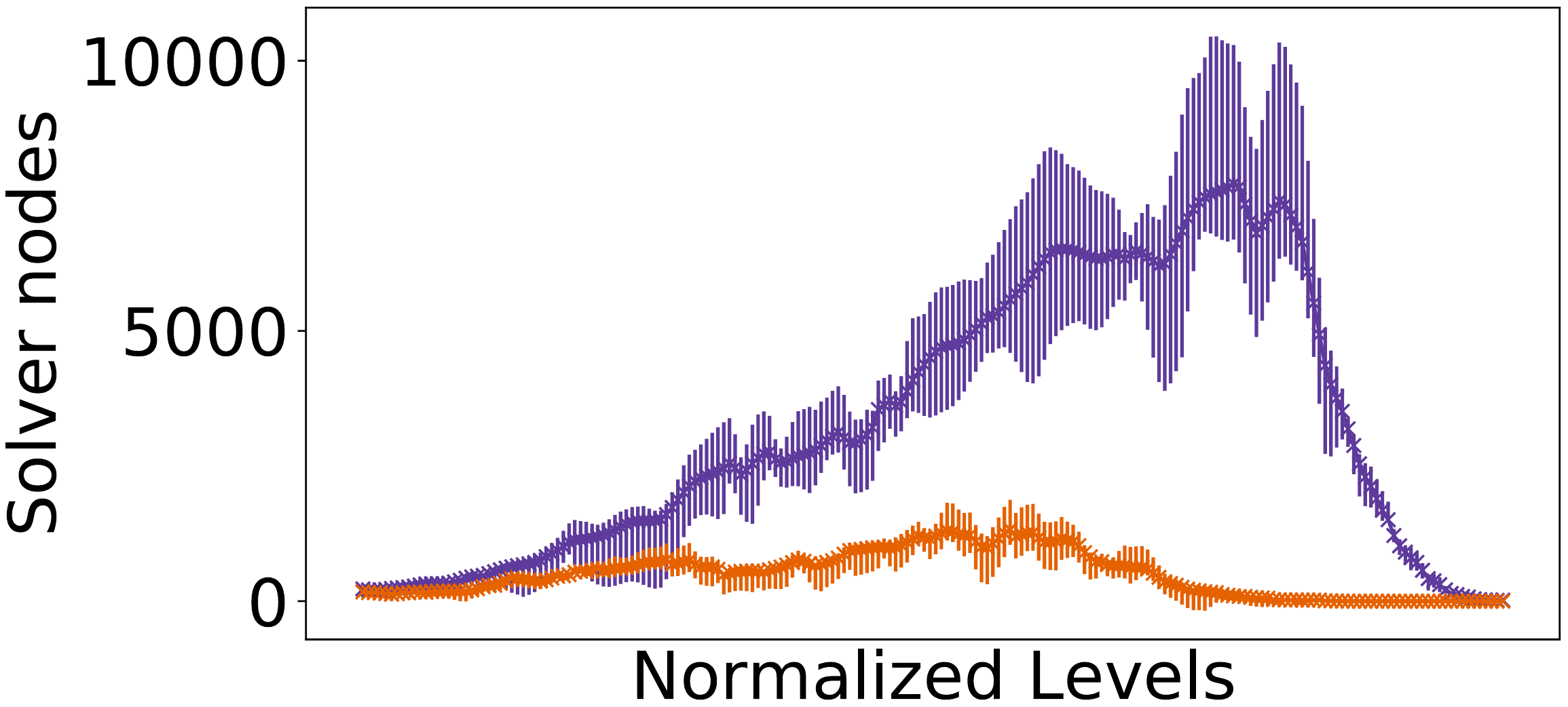}
        \caption{\closed\label{fig:interpol_nodes_closed}}
    \end{subfigure}
    \begin{subfigure}{.5\textwidth}
        \includegraphics[width=\textwidth]{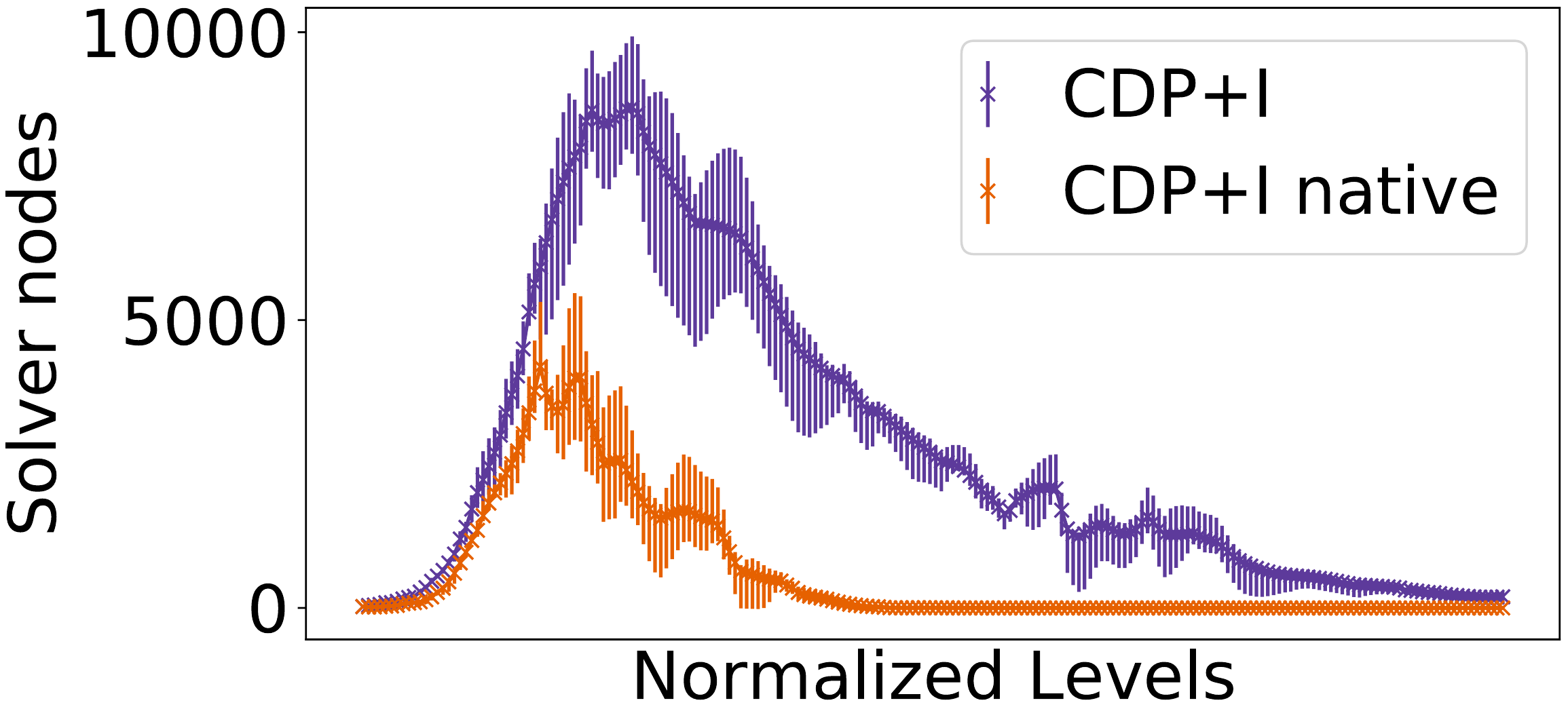}
        \caption{\generator\label{fig:interpol_nodes_gen}}
    \end{subfigure}
    \begin{subfigure}{.5\textwidth}
        \includegraphics[width=\textwidth]{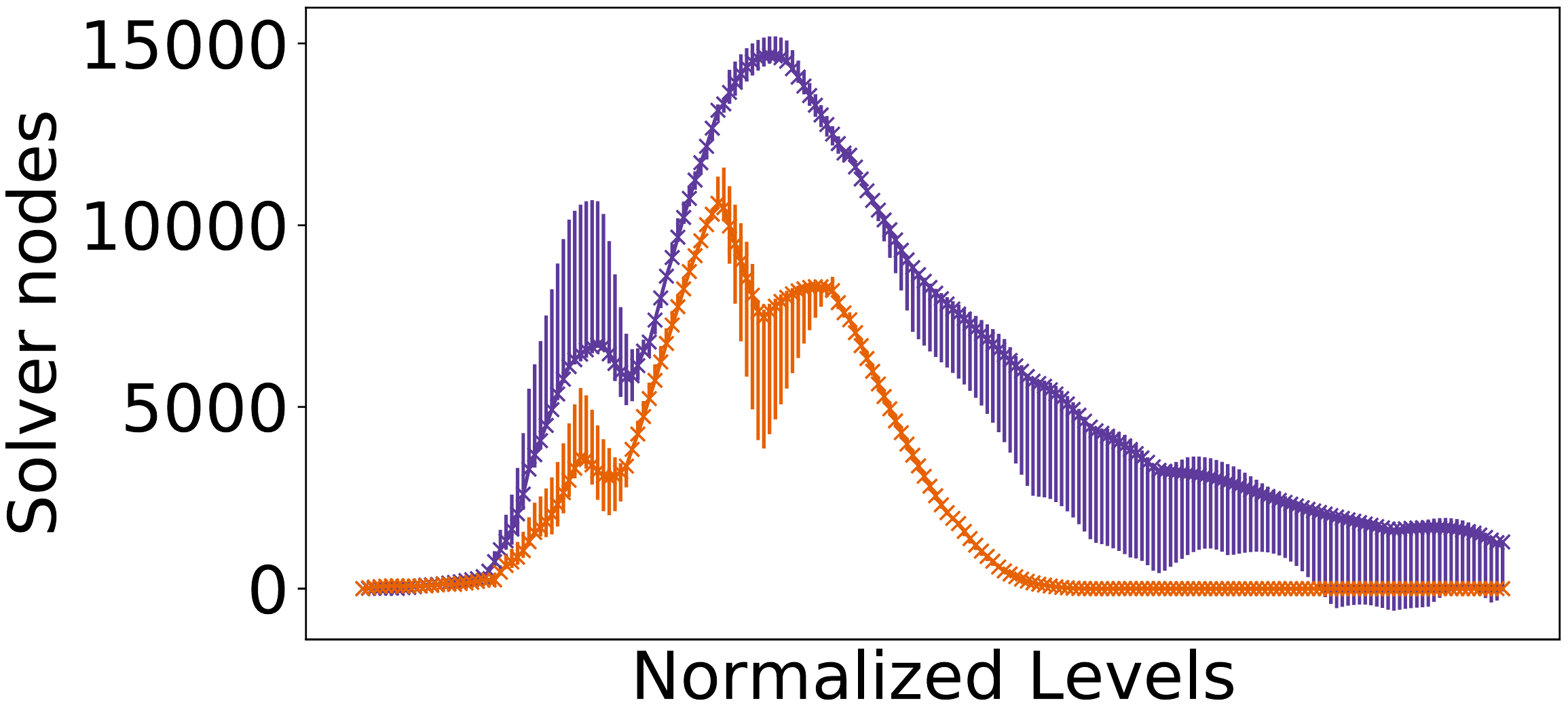}
        \caption{\minimal\label{fig:interpol_nodes_min}}
    \end{subfigure}
    \begin{subfigure}{.5\textwidth}
        \includegraphics[width=\textwidth]{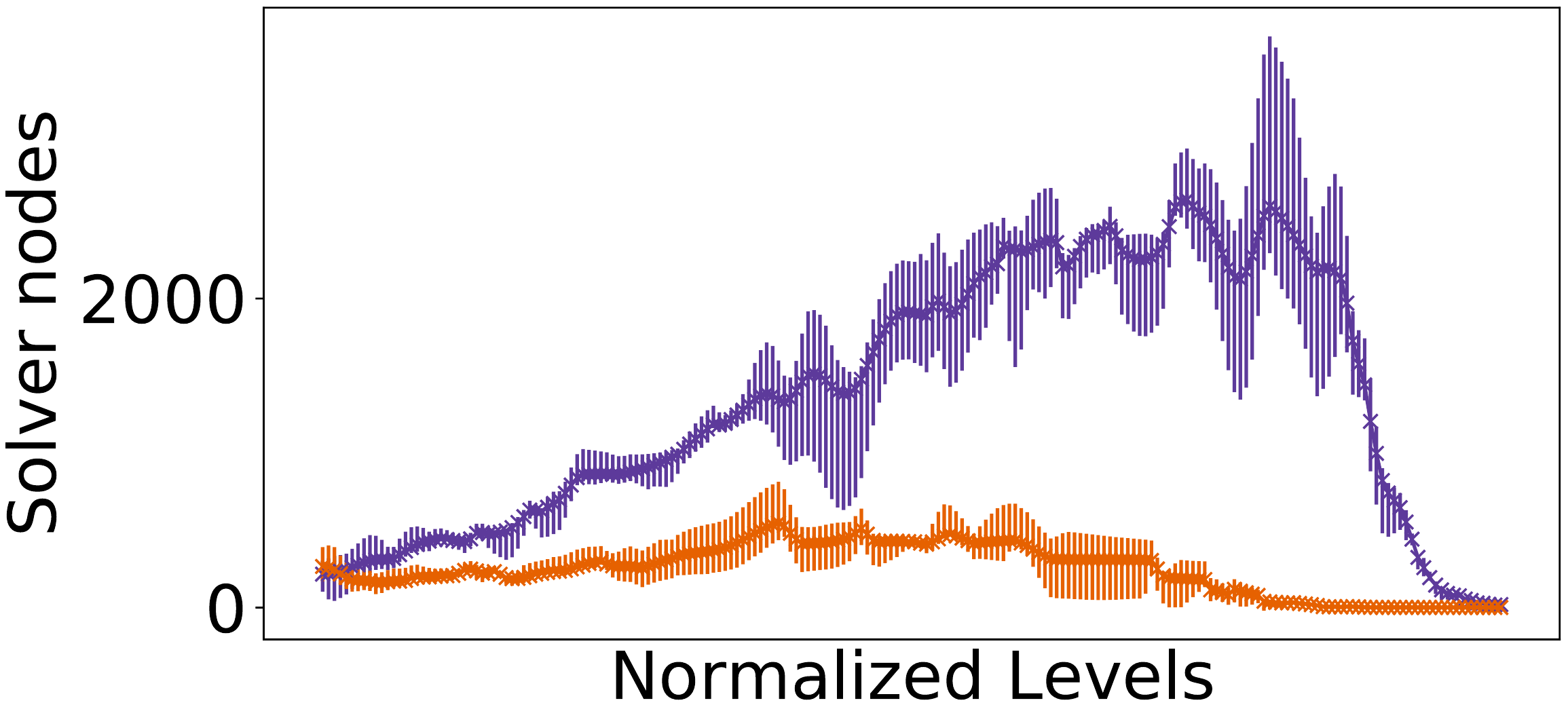}
        \caption{\discriminative\label{fig:interpol_nodes_disc}}
    \end{subfigure}
    \begin{subfigure}{.5\textwidth}
        \includegraphics[width=\textwidth]{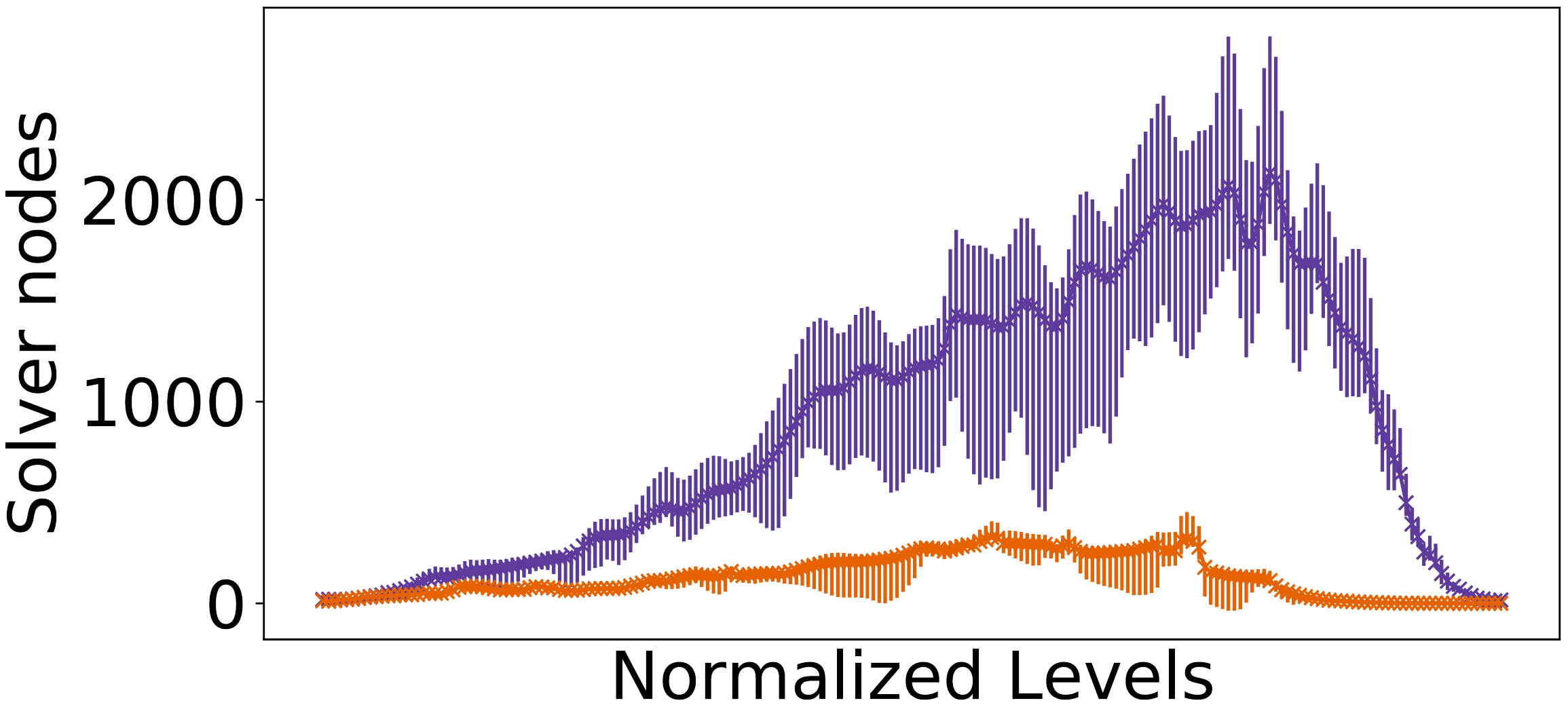}
        \caption{\relevantsub\label{fig:interpol_nodes_rel}}
    \end{subfigure}
    \begin{subfigure}{.5\textwidth}
        \includegraphics[width=\textwidth]{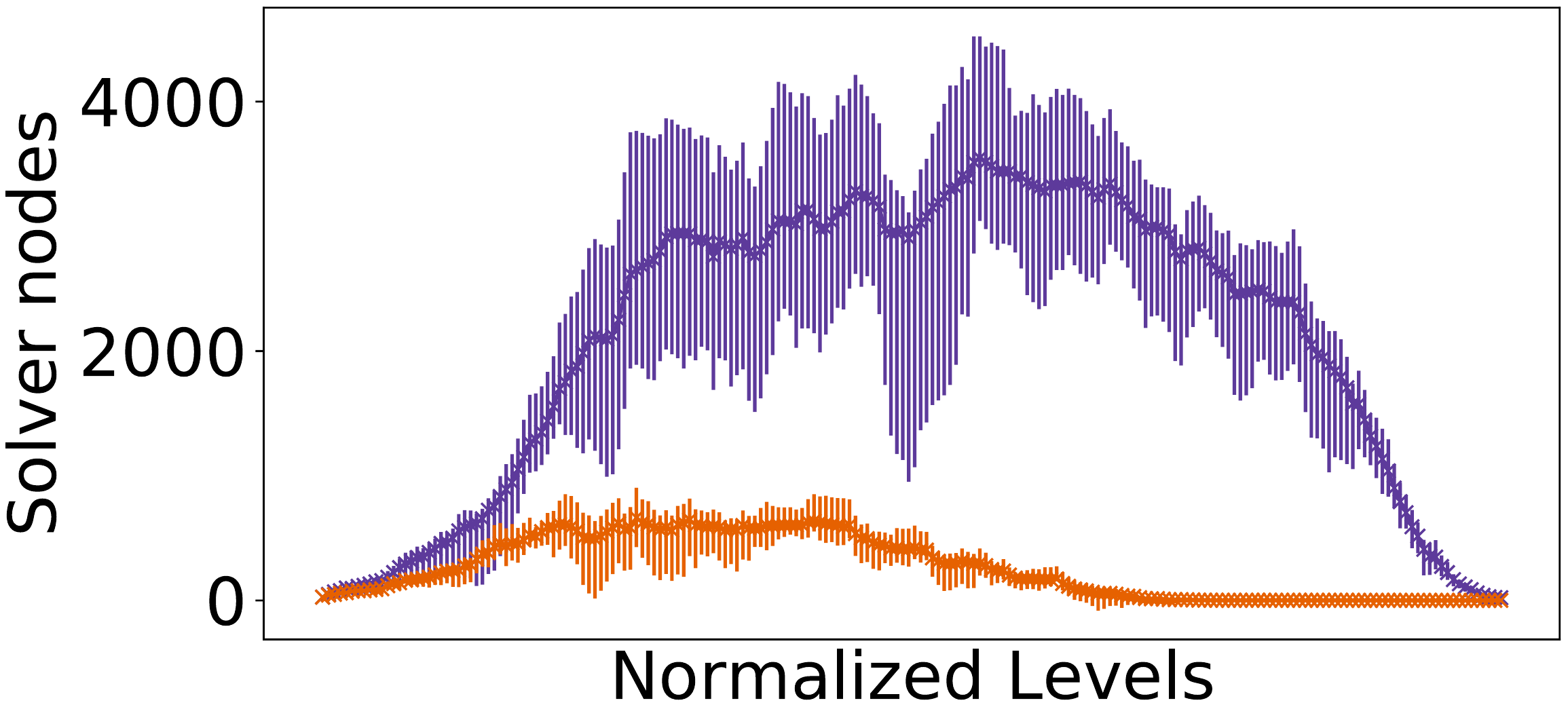}
        \caption{All problem classes\label{fig:interpol_nodes_all}}
    \end{subfigure}
    \caption{Median solver nodes per CDP+I level. Error bars range between the 45\textsuperscript{th} and the 55\textsuperscript{th} percentile. Horizontal axis represents normalised levels between instances. Native CDP+I uses significantly fewer search nodes, thanks to accumulated learned clauses between levels.
    \label{fig:interpol_nodes}}
\end{figure*}

\Cref{fig:interpol_nodes} presents the median number of search nodes per level. Since instances have different numbers of levels, we normalise the number of levels on the horizontal axis. The plot also shows that the default CDP+I's performance can vary amongst different instances, while CDP+I-native's performance has more stability, indicating that CDP+I-native is more robust.

CDP+I-native uses fewer search nodes than pure CDP+I, due to maintaining a subset of learned clauses between levels. \Cref{fig:inter_solve_all} presents a comparison of total solver run time of the two CDP+I variants on \nbc and shows that native interaction clearly results in faster run times as well. 
On PAR2 average, CDP+I-native spends 493 seconds per instance whereas pure CDP+I spends 8,210 seconds.

\subsubsection{A Case Study on \closed Tumor 20\% instance}
To evaluate whether keeping learned clauses improves efficiency, we will demonstrate this by examining one particular instance in detail as a case study.

\begin{figure*}[h!]
    \centering
    \includegraphics[width=0.49\textwidth]{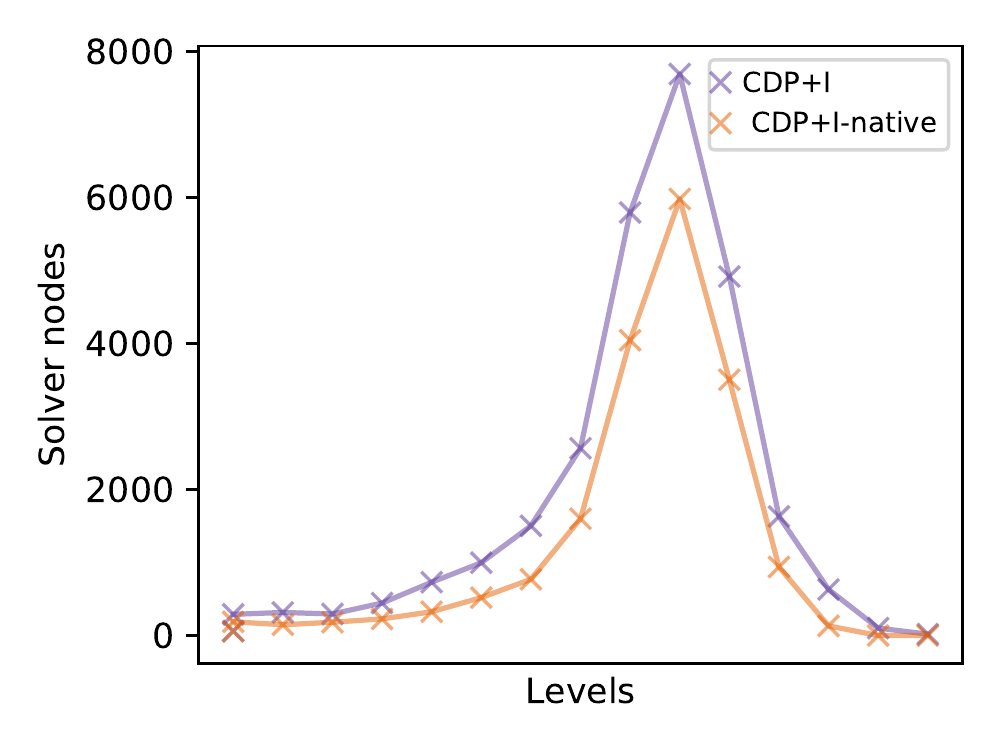}
    \includegraphics[width=0.49\textwidth]{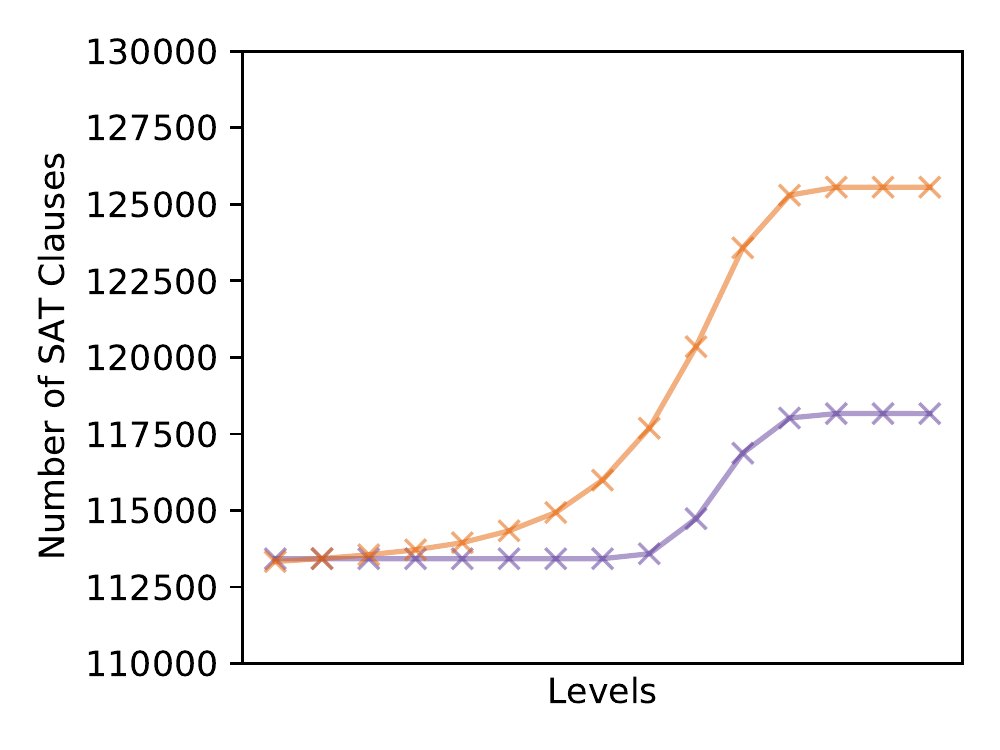}
    \caption{ A comparison on one CDP+I instance with and without native interaction using \nbc{} AllSAT solver. The example instance is \closed{} Tumor with 20\% frequency. Each plot is averaged out from a single model and multiple random seeds. The plot on the left shows the number of solver nodes on each level, while the plot on the right shows the total number of SAT clauses on each level.}
    \label{fig:closed_tumor_20_nodes_satc}
\end{figure*}

\Cref{fig:closed_tumor_20_nodes_satc} presents two plots. The first shows that CDP+I-native uses fewer search nodes on each level. The second illustrates the increased number of SAT clauses in each level that result from keeping learnt clauses. The improved efficiency seen on the first plot is a direct result of the restricted search space from having more clauses. 

\subsubsection{Computational Evaluation with a Standard SAT Solver}

CDP+I on a standard SAT solver operates by generating solution blocking clauses between each solver call in a level. Once a level is completed, the dominance blocking clauses generated by \savilerow are encoded and passed on to the next level. The solution blocking clauses are not encoded again since they are redundant and already implied in the dominance blocking constraints.

Implementing a native interactive system on a standard SAT solver will bring both costs and benefits to its performance. AllSAT solvers are already capable of keeping learned information in a level due to their all solution enumeration behaviour. The native interaction will grant the standard SAT solver this capability, in addition to making the learned information persistent between levels. Thus, the increase of the standard SAT solver's performance will be relatively much higher than the increase of the AllSAT solver's performance. However, since we will still be using solution blocking clauses in a level and since the system cannot eliminate the redundant solution blocking clauses once the level is done, the standard SAT model might expand far beyond its non-native equivalent. AllSAT solvers are not susceptible to this because they can operate without the use of solution blocking clauses, regardless of whether they use native interaction.

\begin{figure*}[h!]
    \begin{subfigure}{.5\textwidth}
        \centering
        \includegraphics[width=0.99\textwidth]{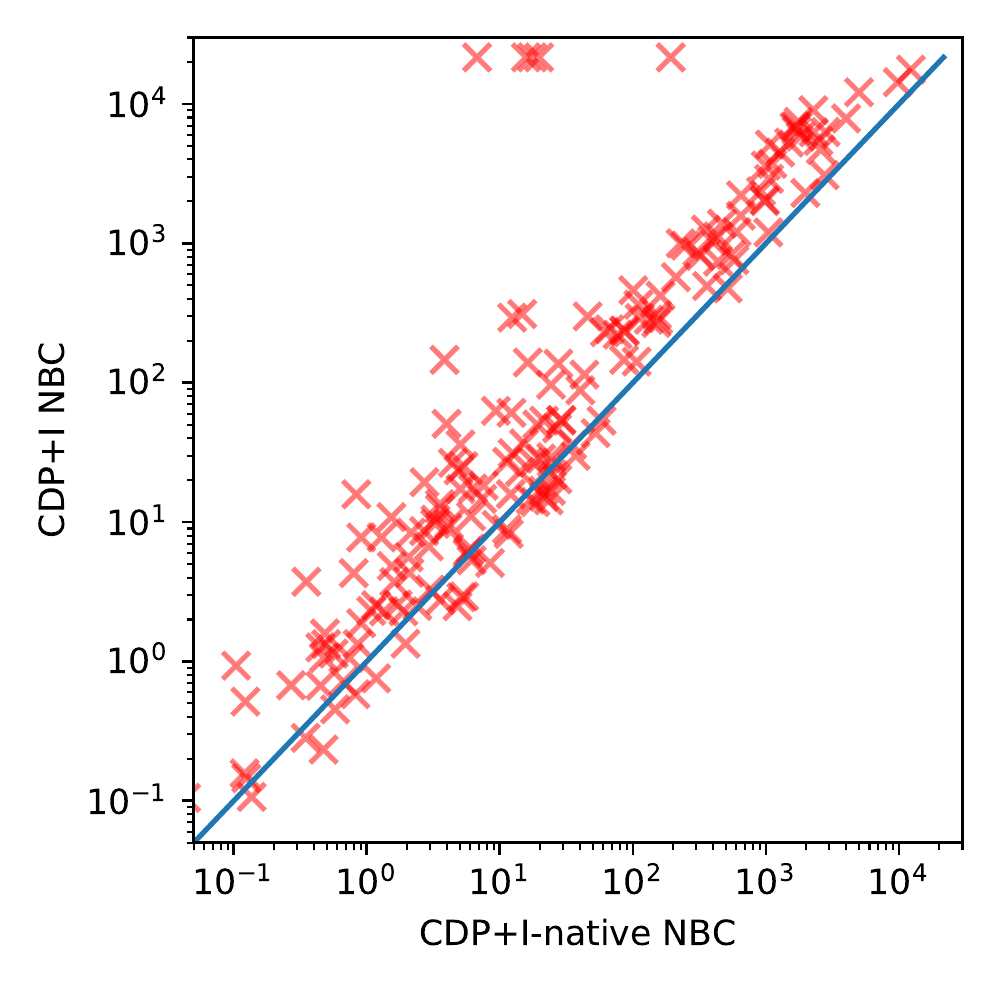}
        \caption{Comparing total solver time using the AllSAT solver \nbc.
        \label{fig:inter_solve_all}}
    \end{subfigure}
    \begin{subfigure}{.5\textwidth}
        \centering
        \includegraphics[width=0.99\textwidth]{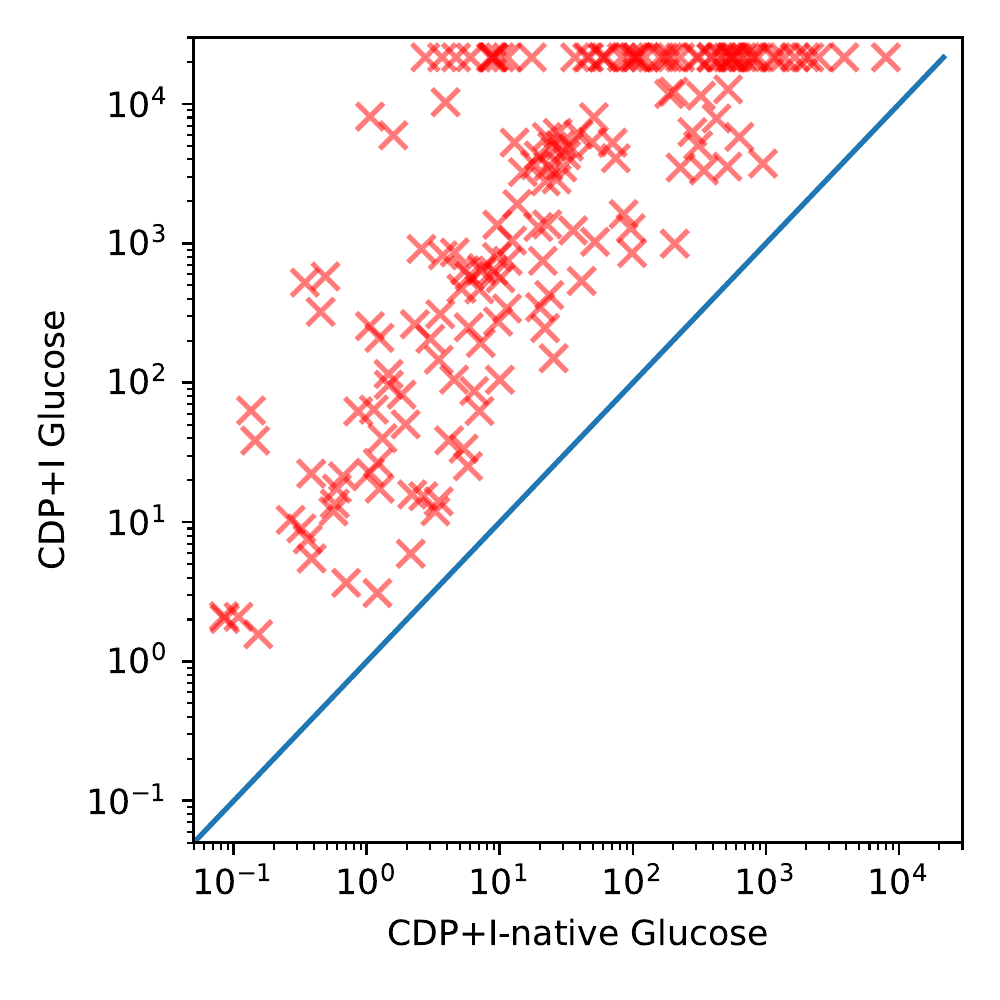}
        \caption{Comparing total solver time using the standard SAT solver \glucose.
        \label{fig:inter_solve_all_glucose}}
    \end{subfigure}
    \caption{Comparison plot between pure CDP+I and CDP+I-native. The time limit is 6 hours per instance. Each data point is averaged out from a single model and multiple random seeds.}
\end{figure*}

\Cref{fig:inter_solve_all_glucose} illustrates a comparison of CDP+I with and without native interaction using the standard SAT solver \glucose. Native interaction increases the performance amongst all instances significantly. The results also suggest that the anticipated decrease in performance due to the expansion of the model did not outweigh the increase provided by native interaction.

In this section we have evaluated the effect of native interaction on the performance of CDP+I. We conducted our analysis on an AllSAT solver and a standard SAT solver. In the next section we evaluate the configuration space of CDP+I-native.

\section{Conclusion}

We have proposed and implemented a new native interaction component to bridge the gap between low level SAT solving and higher level model compilation in \savilerow. We integrated this component into \savilerow to be able to use in the CDP+I framework and optimization problems. Our experiments on different pattern mining tasks and an optimization problem (MRCPSP) show that the native component boosted solving performance significantly. This interaction enabled accessing SAT assumptions to encode level information in a transparent way and also made learned information persistent across multiple runs.

Future work includes evaluating the native interaction component with different problem classes.
We believe this native interaction can be a viable option for multi objective optimization tasks as well. Additionally, there is a large space of possible configurable options which is yet to cover, including different modelling and reformulation methods, other SAT solvers and SMT solvers. 

\paragraph{Acknowledgements} This work is supported by EPSRC grant EP/P015638/1. Nguyen Dang is a Leverhulme Trust Early Career Fellow (ECF-2020-168). 

\bibliographystyle{splncs04}
\bibliography{references}

\clearpage
\appendix
\section{Essence specification for MRCPSP}
\label{sec:ap_mrcpsp}

\begin{figure}[h!]
\centering
\linespread{1.3}
\begin{lstlisting}[style=essence, breaklines]
language Essence 1.3
given nonRenewableResources new type enum
given renewableResources new type enum
given jobs new type enum
given startDummy, endDummy : jobs
given modes new type enum
given renewableLimits: function (total) renewableResources --> int
given nonRenewableLimits  : function (total) nonRenewableResources --> int
given successors : function (total) jobs --> set of jobs
given renewableResourceUsage : 
    function  (jobs, modes, renewableResources) --> int
given nonRenewableResourceUsage : 
    function (jobs, modes, nonRenewableResources) --> int
given duration : function  (jobs,modes) --> int
given horizon : int
letting timesRange be domain int(1..horizon)
find start: function (total)  jobs --> timesRange
find mode: function (total) jobs --> modes
find jobActive: function (total) (jobs,timesRange) --> bool
such that
forAll job : jobs .
    forAll jobSuccessor in successors(job) .
        start(jobSuccessor) >= start(job) + duration((job,mode(job)))
such that
forAll job : jobs .
    forAll time : timesRange .
        jobActive((job,time)) <->(
            time >= start(job) /\ time < start(job) + duration((job,mode(job))))
such that
forAll resource : nonRenewableResources .
    sum([nonRenewableResourceUsage((job, mode(job), resource) )| job : jobs])
        <= nonRenewableLimits(resource)
such that
forAll resource : renewableResources .
    forAll time : timesRange .
        sum([renewableResourceUsage((job,mode(job),resource)) | 
            job : jobs, jobActive((job,time))])
                <= renewableLimits(resource)
such that
start(startDummy)=1
minimising start(endDummy)
\end{lstlisting}
\caption{Essence specificaton for MRCPSP}
\label{fig:mrcpsp}
\end{figure}

\end{document}